\documentclass[letterpaper, 10 pt, conference]{ieeeconf}  % Comment this line out if you need a4paper

\IEEEoverridecommandlockouts                              % This command is only needed if 
\usepackage{cite}
\usepackage{amsmath,amssymb,amsfonts}
\usepackage{graphicx}
\usepackage{textcomp}
\usepackage{xcolor}
\usepackage{blindtext}% for dummy text only
\usepackage{graphicx}
\usepackage{bm}
\usepackage{lipsum}
\usepackage{mwe}% http://ctan.org/pkg/mwe
\usepackage{babel}
\usepackage{cuted}
\usepackage{url}
\usepackage{hyperref}
\hypersetup{
    colorlinks=true,
    linkcolor=black,
    filecolor=magenta,    
    citecolor=black,
    urlcolor=blue,
    }
\usepackage{algorithm}
\usepackage{algorithmic}

\newtheorem{definition}{Definition}
\newtheorem{lemma}{Lemma}

\newtheorem{remark}{Remark}

\graphicspath{{images/}}
\allowdisplaybreaks

\def\be{\boldsymbol e}

\def\bu{\boldsymbol u}
\def\bv{\boldsymbol v}

\def\bp{\boldsymbol p}

\def\R{\mathbb R}
\def\Z{\mathbb Z}
\def\h{\mathrm h}

\def\P{\mathbf P}
\newcommand{\V}{\mathbf V}
\newcommand{\E}{\mathbf E}
\newcommand{\Var}{\mathbf{Var}}
\newcommand{\trace}{\mathbf{trace}}

\def\Log{\mathsf{Log}}
\def\Exp{\mathsf{Exp}}

\def\bbu{\bar{\bm{u}}}
\def\bq{{\bm{q}}}
\def\bzeta{\bm{\zeta}}

\title{\LARGE \bf
Underwater target 6D State Estimation via UUV Attitude Enhance Observability}

\author{Fen Liu,~Chengfeng Jia,~Na Zhang,~Shenghai Yuan,~\textit{Member,~IEEE},\\~Rong Su,~\textit{Senior Member,~IEEE}
 \thanks{The work is supported by National Research Foundation of Singapore under its Medium-Sized Center for Advanced Robotics Technology Innovation and by Naval Group Far East Pte Ltd via an RCA with NTU.}% <-this % stops a space
\thanks{F. Liu, C. Jia, N. Zhang, S. Yuan, R. Su are with the School of Electrical and Electronic Engineering, Nanyang Technological University, Singapore 639798, Singapore (e-mail: {fen.liu, chengfeng.jia, zhan0680, shyuan, RSu}@ntu.edu.sg).}
}

\begin{document}

% \vspace{-4cm}
% \let\oldtwocolumn\twocolumn
% \renewcommand\twocolumn[1][]{%
%     \oldtwocolumn[{#1}{
%     \begin{center}
%           \includegraphics[width=\textwidth]{AS.pdf}
%           \captionof{figure}{Overall illustration of proposed work.}
%           \label{as}
%         \end{center}
%     }]
% }

 \maketitle

% \begin{strip}
% \begin{minipage}{\textwidth}\centering
% \vspace{-90pt}
% \centering
% \includegraphics[width=1\textwidth]{overarching.pdf}
% %\vspace{-8pt}
% \captionof{figure}{Illustrations of the drone swarm overwatch concepts for convoy escort of target 1.}
% \label{overarching}
% \vspace{-10pt}
% \end{minipage}
% \end{strip}

\thispagestyle{plain}
\pagestyle{plain}

%%%%%%%%%%%%%%%%%%%%%%%%%%%%%%%%%%%%%%%%%%%%%%%%%%%%%%%%%%%%%%%%%%%%%%%%%%%%%%%%
\begin{abstract}
Accurate relative state observation of Unmanned Underwater Vehicles (UUVs) for tracking uncooperative targets remains a significant challenge due to the absence of GPS, complex underwater dynamics, and sensor limitations. Existing localization approaches rely on either global positioning infrastructure or multi-UUV collaboration, both of which are impractical for a single UUV operating in large or unknown environments. To address this, we propose a novel persistent relative 6D state estimation framework that enables a single UUV to estimate its relative motion to a non-cooperative target using only successive noisy range measurements from two monostatic sonar sensors. Our key contribution is an observability-enhanced attitude control strategy, which optimally adjusts the UUV’s orientation to improve the observability of relative state estimation using a Kalman filter, effectively mitigating the impact of sensor noise and drift accumulation. Additionally, we introduce a rigorously proven Lyapunov-based tracking control strategy that guarantees long-term stability by ensuring that the UUV maintains an optimal measurement range, preventing localization errors from diverging over time. Through theoretical analysis and simulations, we demonstrate that our method significantly improves 6D relative state estimation accuracy and robustness compared to conventional approaches. This work provides a scalable, infrastructure-free solution for UUVs tracking uncooperative targets underwater.
\end{abstract}

% %%%%%%%%%%%%%%%%%%%%%%%%%%%%%%%%%%%%%%%%%%%%%%%%%%%%%%%%%%%%%%%%%%%%%%%%%%%%%%%%
 \section{Introduction}
% 2x  none-overlaping single bean sonar range measuement
% relative 6 dof state estimation
% % underwater unmanned vehicle localization and perception are extremely hard. 
% % localization 
% % without global coorindate, do not need anything
% % only need multiple scan from single sensor 

% % existing system either need self global coorindate or need multiple agent to collaborate 
% 
%

%
%
%
%
%
%
%第一段：海底目标追踪很重要，很难
%第二段：获取目标的位置或者相对位置的sensor
%第三段：获取目标的位置或者相对位置的算法
%第四段：我们的这个工作做了什么，贡献是什么
%
The tracking of underwater targets is essential for exploration, environmental monitoring, investigation of marine resources and preservation of marine life \cite{isbitiren2011three,liu2020underwater}. However, achieving accurate tracking in underwater environments is inherently challenging \cite{masmitja2023dynamic}. The absence of GPS makes global localization difficult \cite{yuan2021survey}, while conventional tracking methods often struggle with environmental constraints such as poor visibility \cite{vargas2021robust}, signal attenuation \cite{pal2022communication}, and sensor drift \cite{esfahani2019aboldeepio}. Moreover, targets are typically non-cooperative \cite{liu2023multiple}, providing no active state information \cite{nguyen2019single}, and UUV motion patterns can degrade system observability \cite{duecker2020towards}, further complicating estimation accuracy \cite{joshi2023sm}. These factors underscore the urgent need for more robust tracking algorithms that can enhance observability and improve estimation precision.

Various sensing modalities have been explored for underwater target localization, each with distinct advantages and limitations. Monostatic sonar \cite{petillot2001underwater} is widely used for accurate range measurement due to its affordability and simplicity, but suffers from limited tracking capability due to its narrow sensing beam. Side-scan sonar (SSS) improves spatial awareness, but is primarily used for mapping and cannot provide accurate range or bearing measurements. Advanced sonar technologies such as 2D imaging sonar \cite{munoz2024learning} and 3D multi-beam sonar offer richer spatial information but come with high costs, increased power consumption, and computational complexity. Acoustic positioning systems like Ultra-Short Baseline (USBL) \cite{rypkema2019passive} and Long Baseline (LBL) \cite{chen2015underwater} provide accurate localization but rely on pre-deployed infrastructure, making them less suitable for dynamic environments. Optical sensors \cite{xanthidis2021aquavis} deliver high-resolution imagery but become ineffective in turbid or deep-sea conditions, while Doppler-based sensors, such as Doppler Velocity Logs (DVLs) and Acoustic Doppler Current Profilers (ADCPs), excel in self-localization but are not inherently designed for external target tracking. Therefore, there is a need for a sensing approach that balances cost, deployment feasibility, and tracking accuracy to achieve reliable real-time target localization in diverse underwater environments.

Based on measurement data, traditional state estimation methods, including Kalman Filter (KF) \cite{widy2017robust}, Extended Kalman Filter (EKF) \cite{potokar2021invariant}, Particle Filter (PF) \cite{maurelli2022auv}, and Probability Data Association (PDA) \cite{doherty2019multimodal}, rely heavily on sensor measurements but suffer performance degradation in high-noise environments and under sparse observations, as they assume well-characterized system dynamics and observation models. More recently, deep learning-based approaches have been explored for underwater target tracking, leveraging end-to-end feature learning to extract spatial and temporal patterns from raw sensor data. Although these methods demonstrate strong predictive capabilities, they require large labeled datasets, which are often difficult to obtain in underwater scenarios, and their high computational cost limits real-time applicability in resource-constrained UUV systems. Meanwhile, reinforcement learning (RL) \cite{masmitja2023dynamic} has shown promise in adaptive decision-making for UUV path planning, enabling agents to learn optimal tracking strategies through interaction with the environment. However, its effectiveness in partially observable and uncertain underwater conditions remains uncertain, as RL-based models typically require extensive training in simulated environments and may struggle with real-world generalization due to domain discrepancies and limited sensory information. Therefore, a robust and efficient state estimation framework is required to handle sparse and noisy measurements.

%To address these challenges, we propose a sensor fusion solution consisting of an SSS and two monostatic sonars for target ranging and tracking. The fusion solution is achieved by a 6D relative state estimation algorithm designed for autonomous unmanned underwater vehicle (UUV) exploration and target tracking, relying exclusively on range measurements. Given that a single 1D range measurement is insufficient for precise localization in 3D space, incorporating multiple distance measurements introduces higher sensor costs and increased measurement noise. According to distance-rigidity theory, accurate localization in 2D space requires range data from at least three non-collinear anchors to determine three distances, while in 3D space, at least four non-coplanar anchors are necessary to establish four distances \cite{jiang2016simultaneous,cano2023ranging}. However, it is true only when there is enough measurement in view to form such observations. Our approach provides a cost-number balance solution that remains robust in high-noise and communication-free environments. The main contributions are as follows:
To address these challenges, leveraging the sensor fusion of SSS and two monostatic sonars for target detection and ranging, this study proposes a 6D relative state estimation algorithm for autonomous UUV exploration and target tracking based only on noisy range measurements. Given that a single range measurement is insufficient for localization in 3D space, while obtaining multiple distances increases both sensor cost and measurement noise. According to distance-rigidity theory, accurately localizing a target in 2D space requires measurements from at least three non-collinear anchors to determine three distances, while in 3D space, it demands at least four non-coplanar anchors to determine four distances \cite{jiang2016simultaneous,cano2023ranging}. Our approach provides a cost-number balance solution that remains robust in high-noise and communication-free environments. The main contributions are as follows:

Firstly, unlike traditional methods that passively rely on sufficient sensor data \cite{fang2020graph,zhao2016localizability,liu2023relative}, our approach utilizes only monostatic sonar measurements to obtain two noisy distance readings, which are then transformed into linearized measurements of the 6D relative state to the target, even under unknown target motion. This method does not require the UUV to acquire its global position or rely on external communication, making it a practical alternative for real-world underwater applications.

Secondly, unlike conventional approaches that depend on trajectory-motion-compensated observations \cite{hungcooperative,nguyen2019persistently,li2022three,he2019trajectory}, the proposed method actively adjusts the attitude of UUV to ensure that measurements satisfy the Persistent Excitation (PE) condition. This guarantees uniform observability, with Kalman Filtering (KF) applied for state estimation.

\section{Preliminaries and Problem definition}
\subsection{System model} \label{System_model}
This work focuses on how a UUV persistently estimates the relative state between itself and the target, as the target is uncooperative and cannot communicate with the UUV. Let $b$ be the body frame of the UUV at instant $k$, and $w$ be the world frame. A rotation matrix from frame $b$ to frame $w$ is defined as $\bm{R}\in SO(3)$, where $SO(3):=\{\bm{R}\in\R^{3\times3}|\bm{R}^\top \bm{R}=I,\det\{\bm{R}\}=1\}$.

For UUV, we define $\bm{X} \triangleq ({\bp},{\bv})\in \R^{6}$ as its body state vector, where ${\bp} \in \R^3$ is the position, ${\bv} \in \R^3 $ is the velocity, w.r.t world frame $w$. The following kinematic model is considered \cite{forster2015imu}:
\begin{align}
\bm{X}(k+1)=&A\bm{X}(k)+B \bm{R}(k)\cdot{^{b}\bu}_1(k),\label{eq:1}\\
\bm{R}(k+1)=&\Exp\big({}^{b}\bu_2(k)t\big) \bm{R}(k),\label{eq:2}
\end{align}
where $A=
\begin{bmatrix}
    I      &tI \\
    \bm{0} & I \\
\end{bmatrix}\in\R^{6\times6}$, $
B=
\begin{bmatrix}
0.5t^2I\\
tI  \\
\end{bmatrix}\in \R^{6\times3}$, and $t$ is the sampling period. ${^{b}\bu}_1 =[^{b}\bu_{x},^{b}\bu_{y},^{b}\bu_{z}]^\top\in\R^3$ and ${^{b}\bu}_2=[^{b}\bu_{\phi},^{b}\bu_{\theta},^{b}\bu_{\psi}]^\top\in\R^3$ are the linear acceleration and the angle velocity, w.r.t body frame $b$, respectively.

Similarly, the state of the target is denoted by ${\bar{\bm{X}}} \triangleq ({\bar{\bp}},{\bar{\bv}})\in \R^{6}$ with the position ${\bar{\bp}}$ and the velocity ${\bar{\bv}}$, and the initial velocity ${\bv}(0)$ is assumed to be hight. 
Furthermore, we can define the relative motion state between the body of UUV $i$ and target as ${\Upsilon}\triangleq({\bq},{\bm{\vartheta}}) \in \R^6$, where ${\bq}={\bp}-{\bar{\bp}}$ and ${\bm{\vartheta}}={\bv}-{\bar{\bv}}$ are the relative position and relative velocity, respectively. Assume that the target motion acceleration ${\bbu}\in \R^3$ satisfies Gaussian distribution, i.e. $\mathcal{N}(0,W)$ with mean zero and variance $W \in \R^{3 \times 3}$. Here, there exist $\hat{\sigma}\in \R^+$ and $\check{\sigma}\in \R^+$ such that $\hat{\sigma}I_{3\times3}\leq W\leq\check{\sigma}I_{3\times3}$. Then, based on the model \eqref{eq:2}, the dynamics of the relative state ${\Upsilon}$ can therefore be stated as follows:
\begin{equation}\label{eq:3}
\begin{split}
{\Upsilon}(k+1)=&A{\Upsilon}(k)+B(\bm{R}(k)\cdot{^{b}\bu_1}(k)-{\bbu}(k)).
\end{split}
\end{equation}

\subsection{Linearization of range measurements} \label{sec:measurement}
\begin{figure}
\centering
  \includegraphics[width=6cm]{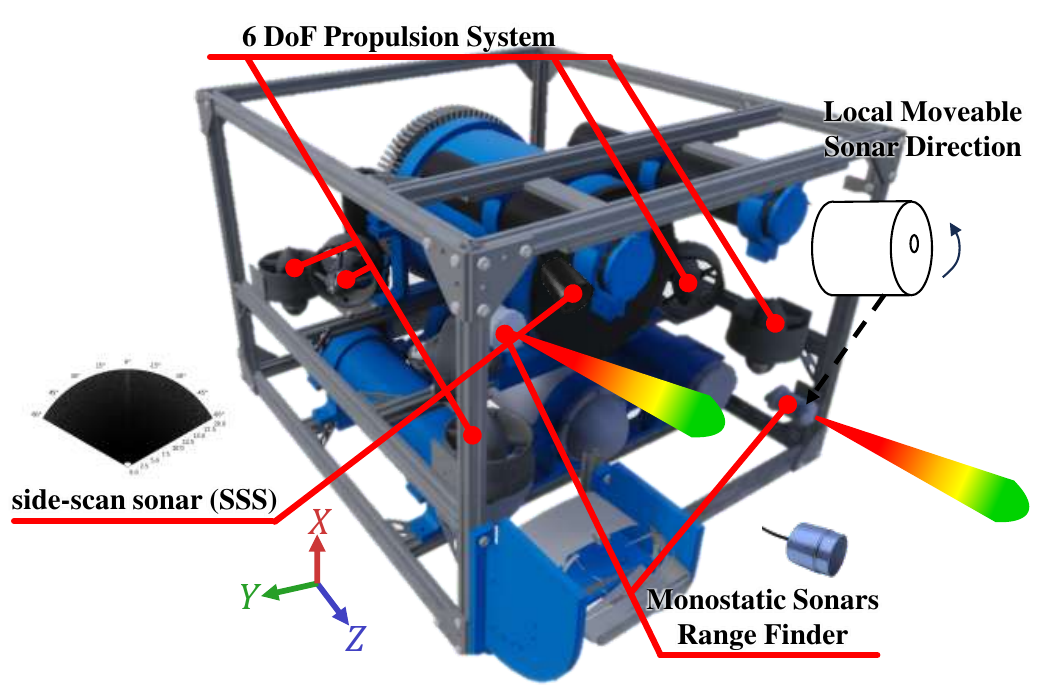}
 \caption{Illustration of UUV system configuration}
  \label{fig:uuv}
\end{figure}

Here, we assume that the target is always within the UUV's sensor measurement range and can be detected by one SSS. As shown in Fig. \ref{fig:uuv}, the UUV is equipped with two monostatic sonars with Pan-Tilt mechanisms for precise range measurement, defined as $s1$ and $s2$.  The typical bandwidth range of a monostatic sonar is 500 Hz to 10 kHz, allowing it to obtain tens to thousands of range points at a single sampling instant. Therefore, we assume that monostatic sonar can acquire $f$ range measurements, i.e. $\Phi_d\triangleq\{d_{i}^{1},d_{i}^{2},\ldots, d_{i}^{f}\}$, at each sampling point $k$. We select one of the measurements for subsequent range linearization, i.e. $d_{i}\in \Phi_d $. Additionally, compared to multistatic sonar, monostatic sonar provides more accurate range measurement information but is still affected by a certain degree of measurement noise, denoted as $v_i(k), i\in\{s1,s2\}$ with zero mean and positive variance $\eta_i$. 
The true distance is defined as $\bar{d}_{i}$, thereby $d_{i}=\bar{d}_{i}+v_i$.

% Here, we assume that the target is always within the UUV's sensor measurement range and can be identified and detected. As shown in the figure, the UUV is equipped with two Monostatic sonars with Pan-Tilt Mechanis for precise range measurement, defined as $s1$ and $s2$. The sensors measure the distances $d_{s1}$ and $d_{s2}$ to the target. Compared to multistatic sonar, monostatic sonar provides more accurate range measurements with typically smaller ranging errors. Therefore, we neglect measurement noise in the analysis. 
Furthermore, the relative position between two sensors in a UUV can be calibrated in the body frame $b$, i.e. ${^{b}}\bq_{s1,s2}=[\bar{s},0,0]^\top,\bar{s}\in \R^{+}$. Additionally, we calibrate the body position of UUV to be centered between the two sensors, ensuring a balanced reference for measurement and localization.

Based on the above, the relative distance measurements can be linearized to derive a new measurement variable ${Y}$, i.e.
\begin{equation}\label{eq:4}
\begin{split}
    {Y}(k)=&0.5\left((d_{s1})^2-(d_{s2})^2\right)\\
    \triangleq& {C}(k){\Upsilon}(k)+\bar{v}_i,
\end{split}
\end{equation}
where ${C}(k)\triangleq{\bq}_{s1,s2}^\top [I\ \bm{0}] $ and ${\bq}_{s1,s2}=\bm{R}\cdot{^{b}\bq}_{s1,s2}$. The unknown noise can be described as $\bar{v}_i=v_{s1}^2+v_{s2}^2+2v_{s1}\bar{d}_{s1}+2v_{s2}\bar{d}_{s2}$ with mean $m(k)=\eta_{s1}+\eta_{s2}$ and variance $\Gamma(k)=2\eta_{s1}^2+2\eta_{s2}^2+4(\frac{1}{f}\sum_{\varrho=1}^{f}(d_{s1}^{\varrho})^2-\eta_{s1})\eta_{s1}+4(\frac{1}{f}\sum_{\varrho=1}^{f}(d_{s2}^{\varrho})^2-\eta_{s2})\eta_{s2}$.

\begin{remark}
    Defining the sensors position $s1$ and $s2$ as $\bp_{s1}$ and $\bp_{s2}$ in the world frame, respectively, we have $d_{s1}^2=(\bp_{s1})^\top (\bp_{s1})-2(\bp_{s1})^\top({\bar{\bp}})+({\bar{\bp}})^\top({\bar{\bp}})+v_{s1}^2+2v_{s1}d_{s1}$ and $d_{s2}^2=(\bp_{s2})^\top (\bp_{s2})-2(\bp_{s2})^\top({\bar{\bp}})+({\bar{\bp}})^\top({\bar{\bp}})+v_{s2}^2+2v_{s2}d_{s2}$. Furthermore, since ${\bp}-{\bp}_{s2}={\bp}_{s1}-{\bp}$, we can obtain ${Y}$ easily. The detailed derivation of variance $\Gamma(k)$ can be found in \cite{liu2024distance}.
\end{remark}

% \begin{assumption}\label{variance bound}
% The variances of target acceleration $\bbu_j$ and measurement noise $\eta$ are assumed to be positive definite with positive upper and lower bounds, i.e., 
% \begin{align*} 
% \hatwI
% &\leq w\leq\checkwI,
% \\
% \hat{\sigma}\leq&\sigma\leq\check{\sigma}, \ i \in \Phi_N,
% \end{align*}
% where $\hatw$, $\checkw$, $\hat{\sigma}$ and $\check{\sigma}$ all are positive numbers.
% \end{assumption}

\subsection{6D state estimation}
% Here, we assume that the target is always within the UUV's sensor measurement range and can be identified and detected. As shown in the figure, the UUV is equipped with two Monostatic sonars with Pan-Tilt Mechanis for precise range measurement, defined as $s1$ and $s2$.  The typical bandwidth range of monostatic sonar is 500 Hz to 10 kHz, allowing it to obtain tens to thousands of range points at a single sampling instant. Therefore, we assume that monostatic sonar can acquire $f$ range measurements, i.e. $\Phi_d\triangleq\{d_{i}^{1},d_{i}^{2},\ldots, d_{i}^{f}\}$, at each sampling point $k$. We select one of the measurements for subsequent range linearization, i.e. $d_{i}\in \Phi_d $. Additionally, compared to multistatic sonar, monostatic sonar provides more accurate range measurement information but is still affected by a certain degree of measurement noise, denoted as $v_i(k), i\in\{s1,s2\}$ with zero mean and positive variance $\eta_i$. 
% The true distance is defined as $\bar{d}_{i}$, thereby $d_{i}=\bar{d}_{i}+v_i$.
Checking Assumptions or Definitions in \cite{reis2023kalman,li2022three,haring2020stability}, we know that to estimate the system state using a Kalman filter (KF), the system must be observable and controllable, and the variance of system input and output noise must be positive definite.  The positive definite variance $W$ of target acceleration input ${\bbu}$ is defined in section \ref{System_model}. Considering the system matrix $A$ and input matrix $B$ are defined in \eqref{eq:1}, the system can be easily proven to be controllable. 

Furthermore, given that the positions of the two sensors are inherently distinct, we can only ensure that the rank of output matrix ${C}(k)^\top {C}(k)$ is greater than or equal to 1, thereby, the observability Gramian matrix cannot be directly determined to be positive definite, even when the system matrix $A$ is invariant and full rank. Therefore, we first assume that the system being estimated is uniformly observable, i.e. Definition \ref{Uniform observability}, which will be satisfied by attitude control in later sections. 

\begin{definition}
\label{Uniform Observability} \cite{deyst1968conditions} (Uniform observability) \label{Uniform observability}
\textit{The matrix pair $\{A, {C}(k)\}$, where $A, {C}(k)\in \R^{3\times3}$ are system and output matrices, respectively, is uniformly completely observable if $\exists$ $\hat{a}_{S2}, \check{a}_{S2} \in \R^{+}$, $M \in \Z^+$ such that for all $k \geq 0$,
\begin{equation}\label{eq: A C PE}
\begin{split}
\hat{a}_{S2}I\leq\sum^{k+M-1}_{m=k}
&(\mathcal{A}^m)^\top {C}(m)^\top {C}(m)\mathcal{A}^m\leq\check{a}_{S2}I,
\end{split}
\end{equation}
where $\mathcal{A}^m=A^{k+M-1-m}$, and $\mathcal{A}^{k+M-1}=I$.}
\end{definition} 

Then, the 6D state estimator can be obtained based on the KF in \cite{reif1999stochastic} as follows:
\begin{align}
&{\hat{\Upsilon}}(k)
={\hat{\Upsilon}}(k|k-1)+\bm{K}(k)\left[Y(k) -\hat{Y}(k)\right],\label{eq9-0}
\\
&{\hat{\Upsilon}}(k|k-1)
=A{\hat{\Upsilon}}(k-1)+B\bm{R}(k-1){}^b{\bu}(k-1),\label{eq9-1}
\\
&\hat{Y}(k)
= {C}(k){\hat{\Upsilon}}(k|k-1)+m(k),\label{eq9-2}
\\
&\bm{K}(k)
=
\xi(k|k-1){C}(k)^\top \big[{C}(k)\xi(k|k-1){C}(k)^\top \nonumber\\
&~~~~~~~~~~~+\Gamma(k)\big]^{-1},\label{eq9-3}
\\
&\xi(k|k-1)
=Aw(k-1)A^\top +BW B^\top,\label{eq9-4}
\end{align}
where ${\hat{\Upsilon}}(k)\triangleq({\hat{\bq}}(k),{\hat{\bm{\vartheta}}}(k)) $ is the estimated relative state with the estimated relative position ${\hat{\bq}}(k)$ and relative velocity ${\hat{\bm{\vartheta}}}(k)$. $\bm{K}(k)$ is the estimator gain, $\xi(k) \triangleq \Var\{{^{b}\Upsilon}-{^{b}\hat{\Upsilon}}\}$ is the estimation error variance, and $\xi(k|k-1)\triangleq\Var\{{^{b}\Upsilon}-A{^{b}\hat{\Upsilon}}(k-1)\}$ is the predicted variance. 

\begin{remark}
As is well known, KF is proven to be optimal in the statistical sense for linear systems, and other variants such as EKF \cite{zhang2023toward}, UKF \cite{li2021joint}, and adaptive KF \cite{hoang2024state} are typically employed only when the classical KF cannot be directly applied. In this work, we have come up with an original stochastic linear observation model, and an appropriate control protocol to guarantee observability. 
\end{remark}

\subsection{Problem formulation} \label{sec: problem fomulation}
This work aims to ensure that the UUV can persistently estimate the relative state between itself and the uncooperative target. Therefore, the following goals must be achieved.

Firstly, we must design the UUV's attitude velocity control ${^{b}\bu}_2(k)$ to enhance the target's observability, that is, Definition \ref{Uniform observability} holds. 

Secondly, the tracking control ${^{b}\bu}_1(k)$ must be optimized to keep the target within the UUV measurement range. 

\section{Observability-enhanced controller design}

\subsection{Attitude control} 
In this section, we aim to design attitude control so that the system being estimated is uniformly observable. Recalling ${C}(k)\triangleq{\bq}_{s1,s2}^\top [I\ \bm{0}]$, and based on the Assumption 2 in \cite{balandi2023persistent} and Lemma 2.3 in \cite{hamel2016riccati}, if the formula \eqref{eq: A C PE} holds, then we can reduce to the relative position ${\bq}_{s1,s2}$ is \textbf{persistently exciting} (PE), i.e. $\exists$ $\hat{a}_{\bq}, \check{a}_{\bq} \in \R^{+} $ and $N \in Z^+$ such that for all $k$: 
\begin{equation}\label{eq:10}
\begin{split}
\hat{a}_{\bq}I\leq\sum_{m=k}^{k+N-1}{\bq}_{s1,s2}(m) {\bq}_{s1,s2}^\top(m)\leq\check{a}_{\bq}I.
\end{split}
\end{equation}

Consider ${\bq}_{s1,s2}=\bm{R}\cdot{}^{b}\bq_{s1,s2}$, we have 
\begin{equation}\label{eq:11}
\begin{split}\Lambda1\triangleq&\sum_{m=k}^{k+N-1}{\bq}_{s1,s2}(m) {\bq}_{s1,s2}^\top(m)\\
=&\sum_{m=k}^{k+N-1}\bm{\bm{R}}(m)\cdot{^b\bq}_{s1,s2}{^b\bq}_{s1,s2}^\top\cdot\bm{R}^\top(m).
\end{split}
\end{equation}

\begin{figure}
\centering
  \includegraphics[width=8.5cm]{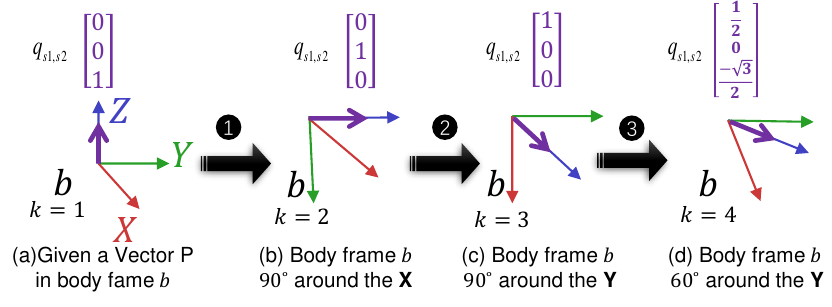}
 \caption{Attitude changes affect relative position in the world frame.}
  \label{fig:rotation}
\end{figure}

If ${^b\bq}_{s1,s2}{^b\bq}_{s1,s2}^\top$ is a non-singular matrix (This can be ensured by allowing the user to select the sensor installation position), we can easily conclude that $\Lambda1$ has positive upper and low bounds for $\bm{R}(k)\bm{R}^\top(k)=I$. Otherwise, we need to design $\bm{R}(k)$ to ensure that ${\bq}_{s1,s2}(k)$ is PE. For instance, we assume 
${^b\bq}_{s1,s2}=[0,0,1]^\top$. As shown in Fig. \ref{fig:rotation}, by rotating the body of the UUV, we can obtain different values of 
${\bq}_{s1,s2}$ in the world frame.

Firstly, we design Algorithm \ref{ALG:The_orientations_algorithm} to make $\bm{R}(k)$ cover a range of diverse orientations. Based on this, ${\bq}_{s1,s2}(k)$ will uniformly distributed over the surface of a sphere with radius $\|{^{b}\bq}_{s1,s2}\|$. Then, we will provide a theoretical feasibility proof in Lemma \ref{lem1: position_Persistently_exciting}.

\begin{algorithm}  
 \caption{Attitude controller 1} 
 \label{ALG:The_orientations_algorithm}
 %\LinesNumbered
    \textbf{1:} Measurement and obtain the initial orientation matrix: $\bm{R}(0)$.\\
    \textbf{2:} Generate a completely random rotation matrix using a random unit quaternion: $\bar{R}(k)$.\\
    \textbf{3:} Obtain attitude controller: ${}^{b}\bu_2(k)=\frac{1}{t}\Log(\bar{R}(k)R(k)^\top)$.\\
    \textbf{4:} Generate new orientation matrix: \eqref{eq:2}.
 \end{algorithm}

%  Therefore, based on the formula \eqref{eq:1}, the sensor angular velocity controller can be designed as
%  \begin{equation}\label{eq:9}
% \begin{split}
% {^{b}\bm{\omega}}(k)=(tJ^{b}(k))^{-1}({^{b}\Phi}(k+1)-{^{b}\Phi}(k)).
% \end{split}
% \end{equation}

\begin{lemma}
\label{lem1: position_Persistently_exciting}
Based on Algorithm \ref{ALG:The_orientations_algorithm}, the relative position ${\bq}_{s1,s2}(k)$ is PE.
\end{lemma}
\textbf{Proof.} Denote on constant $\xi\triangleq\|{^b\bq}_{s1,s2}\|$. According to the properties of the rotation matrix $\bm{R}$, the size of the rotated vector remains unchanged, that is, $\|\bm{R}{^b\bq}_{s1,s2}\|=\xi$. Then, we have $0\leq\lambda(\bm{R}{^b\bq}_{s1,s2}{^b\bq}_{s1,s2}^\top\bm{R}^\top)\leq \xi^2$, thereby $\check{a}_{q}\triangleq\lambda_{\max}\{\Lambda1\}=N\xi^2$. Furthermore, $\bm{R}{^b\bq}_{s1,s2}$ is uniformly distributed on the sphere, which means its direction is isotropic in three-dimensional space. Therefore, we have $\E[\bm{R}{^b\bq}_{s1,s2}]=0$ and $\E[\bm{R}{^b\bq}_{s1,s2}{^b\bq}_{s1,s2}^\top\bm{R}^\top]=\frac{1}{3}\xi^2I_{3}$, where $I_{3}$ is the $3 \times 3$ identity matrix. Then, $\E[\Lambda1]=\frac{N}{3}\xi^2I_{3}$ can be obtained easily. 

Since $\Lambda1$ is a nonnegative definite random variable and based on Chernoff Bounds, there exists $c \in [0,1]$ such that
\begin{equation}\label{eq:12}
\begin{split}
  \P\Big(\lambda_{\min}(\Lambda1)\leq\hat{a}_{q}\big)\leq 3\Big[\frac{e^{-c}}{(1-c)^{1-c}}\big]^{N/3},
\end{split}
\end{equation}
where $\hat{a}_{q}\triangleq(1-c)\frac{N}{3}\xi^2$.

Here, we use the method of contradiction to prove that: $\exists N \in \R^{+}$ such that $\lambda_{\min}(\Lambda1)>\hat{a}_{q}$. Assume that $\forall N \in \R^{+}$ such that $\lambda_{\min}(\Lambda1)\leq\hat{a}_{q}$. As $N\to \infty$, $3\big[\frac{e^{-c}}{(1-c)^{1-c}}\big]^{N/3} \to 0$, which implies that the probability of $\lambda_{\min}(\Lambda1)\leq\hat{a}_{q}$ occurring is zero, thereby contradicting our assumption. Therefore, we have $\lambda_{\min}(\Lambda1)>\hat{a}_{q}>0$ for $\exists c \in [0,1]$ and $\exists N \in \R^{+}$.

Therefore, the relative position ${\bq}_{s1,s2}$ is PE as $\Lambda1$ satisfies condition \eqref{eq:10}.

Although Algorithm 1 enables attitude variation to compensate for the UUV's observation of the target, excessively large attitude changes may cause the target to move out of the UUV's sensor perception range in practical applications, such as target tracking. Therefore, further optimization of attitude control is required.

\begin{figure}
\centering
  \includegraphics[width=6.5cm]{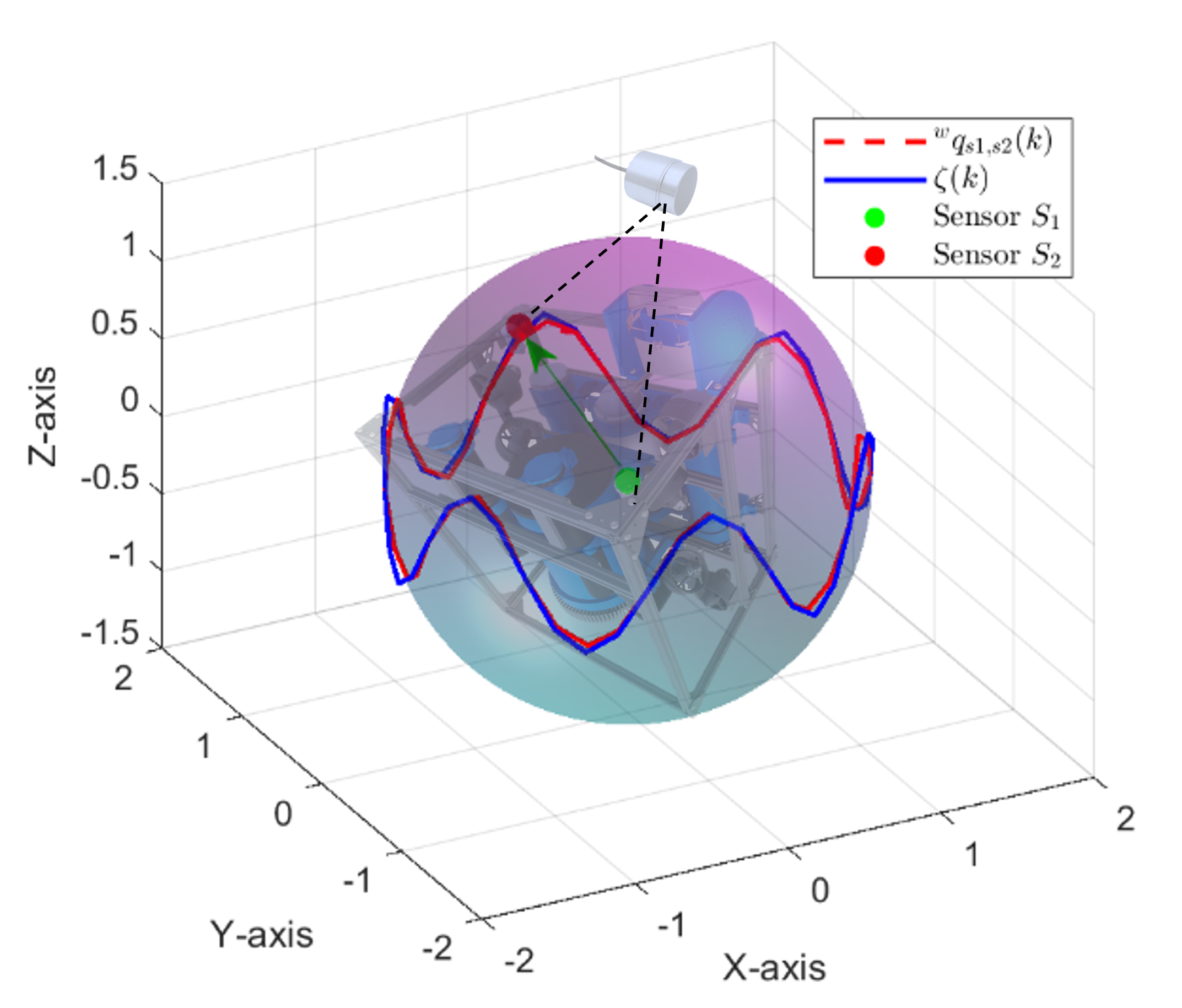}
 \caption{An example of the trajectories of ${\bq}_{s1,s2}(k)$ and $\bzeta(k)$}
  \label{fig:Algorithm_2}
\end{figure}

Secondly, Algorithm 2 is designed so that the vector ${\bq}_{s1,s2}(k)$ can rotate around the XY plane with the position of the sensor $s1$ as the center and oscillate along the Z-axis, i.e.  ${\bq}_{s1,s2}(k)$ satisfies the following function $\bzeta(k)$, 
\begin{equation}
\begin{split}
\bzeta(k)
=
\|{^{b}\bq}_{s1,s2}\|
\begin{bmatrix}
    \sin(\rho k\pi), &\cos(\rho k\pi), &\h(k)
\end{bmatrix}^\top,
\end{split}
\end{equation}
where $\h(k) \in \R$ is the vertical motion function, which can be noise or user-defined, ensuring that $\bzeta(k)$ satisfies the PE condition.  $\rho$ is the angular rotation frequency. Fig. \ref{fig:Algorithm_2} provides an example of the trajectories of ${\bq}_{s1,s2}(k)$ and $\bzeta(k)$.

\begin{algorithm}  
 \caption{Attitude controller 2} 
 \label{ALG:The_orientations_algorithm2}
 %\LinesNumbered
    \textbf{1:} Measurement and obtain the initial orientation matrix: $\bm{R}(0)$, and Give the idea function $\bzeta(k)$.\\
    \textbf{2:} Compute rotation axis $\gamma={^b\bq}_{s1,s2}\bzeta(k)$, and rotation angle $\tilde{\theta}=\arccos(\frac{\gamma}{\|{^b\bq}_{s1,s2}\|\|\bzeta(k)\|})$.\\
    \textbf{3:} Generate target orientation matrix: $\bar{R}(k)=\Exp(\gamma\tilde{\theta})$.\\
    \textbf{4:} Obtain attitude controller: ${}^{b}\bu_2(k))=\frac{1}{t}\Log(\bar{R}(k)\bm{R}(k)^\top)$.\\
    \textbf{5:} Generate new orientation matrix: \eqref{eq:2}.
 \end{algorithm}

\begin{lemma}
\label{lem2: position_Persistently_exciting2}
Based on Algorithm \ref{ALG:The_orientations_algorithm2}, the relative position ${\bq}_{s1,s2}(k)$ is PE.
\end{lemma}
\textbf{Proof.} Based on Algorithm 2, the vector ${\bq}_{s1,s2}(k)$ follows the trajectory of $\bzeta(k)$ on a sphere centered at sensor $s1$ with a radius of ${^{b}\bq}_{s1,s2}$. Since $\bzeta(k)$ satisfies the PE condition, ${\bq}_{s1,s2}(k)$ is also PE.

Based on the above algorithms, when the system being estimated is uniformly observable, i.e. Definition \ref{Uniform observability} holds, in terms of the proof in the previous KF works \cite{liu2024distance}, we conclude that the estimated relative state error in the mean square can converge to a bound $\iota$, i.e. $\iota\geq\E\{\|{\Upsilon}(k)-{\hat{\Upsilon}}(k)\|^2\}$. 

\subsection{Tracking control} \label{Tracking_control}
We assume that the idea related position from the body of UUV to the target is ${}^b\bq^*$ in the body frame. Based on the estimated target state ${\hat{\Upsilon}}(k)$ and our previous works \cite{liu2023non}, the tracking controller can be designed as
\begin{equation}
\begin{split}
{^bu}_{1}(k)=&-\frac{2\alpha}{t^2}(\bm{R}^{-1}(k){\hat{\bq}(k)}-{}^b\bq^*)\\
&-\frac{2}{t}\bm{R}^{-1}(k){\hat{\bm{\vartheta}}}(k),
\end{split}
\end{equation}
where $\alpha$ is used to control the convergence speed.

Then, we analyze the convergence of tracking error under the influence of the aforementioned controller. Defined the tracking error as $\be(k)={\bq(k)}-\bm{R}(k){^bq^*}$ and recalling the system in \eqref{eq:3}, the dynamics of $e(k)$ can be derived as
\begin{equation}
\begin{split}
\be(k+1)=&(1-\alpha)\be(k)+\bar{\alpha}({\Upsilon}(k)-{\hat{\Upsilon}}(k))\\
&+(\bm{R}(k)-\bm{R}(k+1)){^bq^*}-\frac{1}{2}t^2{\bbu}(k)
\end{split}
\end{equation}
Where $\bar{\alpha}=(\alpha I_{3\times3},tI_{3\times3})\in\R^{3\times 6}$.

We select an appropriate Lyapunov function as follows: $\V(k)=\be^\top(k)\be(k)$. Then, considering $\Var\{{\bbu}(k)\}=W$, $\hat{\sigma}I_{3\times3}\leq W\leq\check{\sigma}I_{3\times3}$ defined in \ref{System_model}, $\E\{{\bbu}(k)^\top{\bbu}(k)\}=\trace\{W\}$ and the difference in the mean square can be obtain as 
\begin{equation}
\begin{split}
\E\{\V(k+1)\}=&\E\{\be^\top(k+1)\be(k+1)-\be^\top(k)\be(k)\}\\
&\leq(2(1-\alpha)^2-1)\V(k)+\frac{3}{4}t^4\check{\sigma}+\sigma,
\end{split}
\end{equation}
where $\sigma= \max\big\{2\E\{\|\bar{\alpha}({\Upsilon}(k)-{\hat{\Upsilon}}(k))+(\bm{R}(k)-\bm{R}(k+1)){^bq^*}\|^2\}\big\}$.

Therefore, as long as we choose $\alpha \in (1-\frac{1}{\sqrt{2}}, 1+\frac{1}{\sqrt{2}})$, such that $(2(1-\alpha)^2-1)<0$, we can ensure the convergence of the tracking error $\be(k)$. In other words, under the influence of the tracking controller, the UUV can track the target while maintaining the ideal relative position ${}^b\bq^*$ within a bounded error.

\section{SIMULATIONS AND EXPERIMENTS}
\textbf{Numerical simulation:}
Here, a numerical simulation example is used to demonstrate that the UUV can achieve a persistent estimation of its relative position to the target through attitude control and tracking control. To better align with practical applications, we use only Algorithm 2 for observation compensation.

\begin{figure}
\centering
  \includegraphics[width=8.5cm]{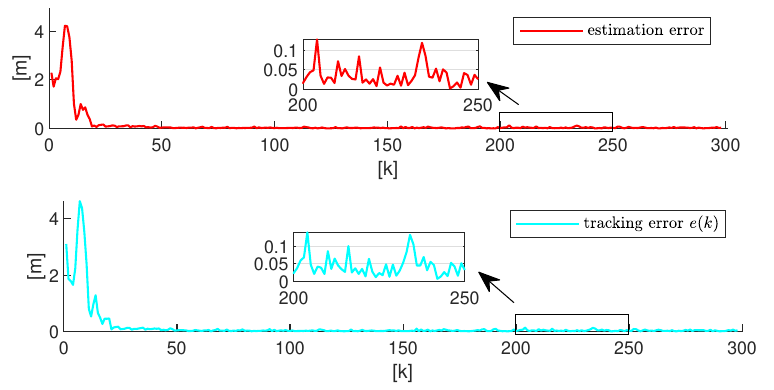}
 \caption{The relative state estimation error trajectory between the UUV and the target, and the UUV’s tracking error trajectory of the target.}
  \label{fig:error}
\end{figure}

\begin{figure}
\centering
  \includegraphics[width=8.5cm]{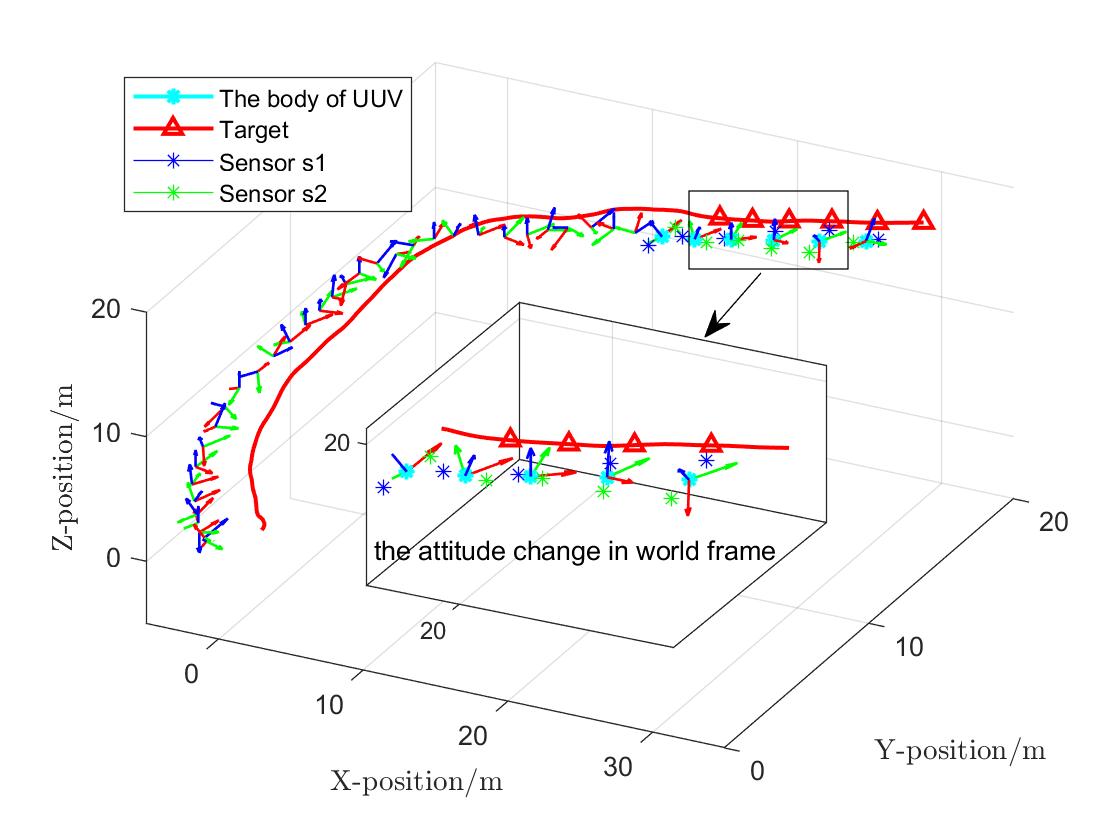}
 \caption{Partial real-time trajectories of the target and UUV, and the real-time attitude changes of UUV.}
  \label{fig:trajectory}
\end{figure}

The acceleration variance of the target is given as $W=[0.004,0,0;0,0.001,0;0,0,0.001]$, and the distance measurement noises variance of two difference sensors are both $v_1=v2=0.01$. The initial values are provided as $t=0.5$, ${\bar{\bp}}(0)=[1;2;2]$, $,{\bar{\bv}}(0)=[0.02;0.1;0.1]$, ${^{b}}\bq_{s1,s2}=[-1;1;0]$, ${\bq^*}=[-2;-2;0]$ and $\bm{R}(0)=I$.The sampling time is $t=0.5$. The variance of target acceleration is given by $W=[0.004,0,0;0,0.001,0;0,0,0.001]$ and the variances of range measurement noise are $\eta_1=\eta_2=0.01$. The ideal motion trajectory of the relative position ${\bq}_{s1,s2}(k)$ is defined as \begin{equation*}
\begin{split}
\bzeta(k)
=
\|{^{b}\bq}_{s1,s2}\|
\begin{bmatrix}
    \sin(\frac{1}{24}k\pi), &\cos(\frac{1}{24}k\pi), &\frac{1}{2}\cos(\frac{1}{6}k\pi)
\end{bmatrix}^\top.
\end{split}
\end{equation*}
In the numerical simulation, the angular rotation frequency of $\bzeta(k)$ is set relatively high because we aim for rapid convergence of the state estimation errors and for them to be as close to zero as possible. However, in real-world applications, the angular rotation frequency should be lower to ensure that UUV attitude changes do not cause the target to move out of the sensor's measurement range.

Moreover, based on the convergence analysis in section \ref{Tracking_control}, the tracking controller gain is given as follows: $\alpha=1.2$. All other quantities that require an initial value are set to zero.

\begin{figure*}
\centering
  \includegraphics[width=1\linewidth]{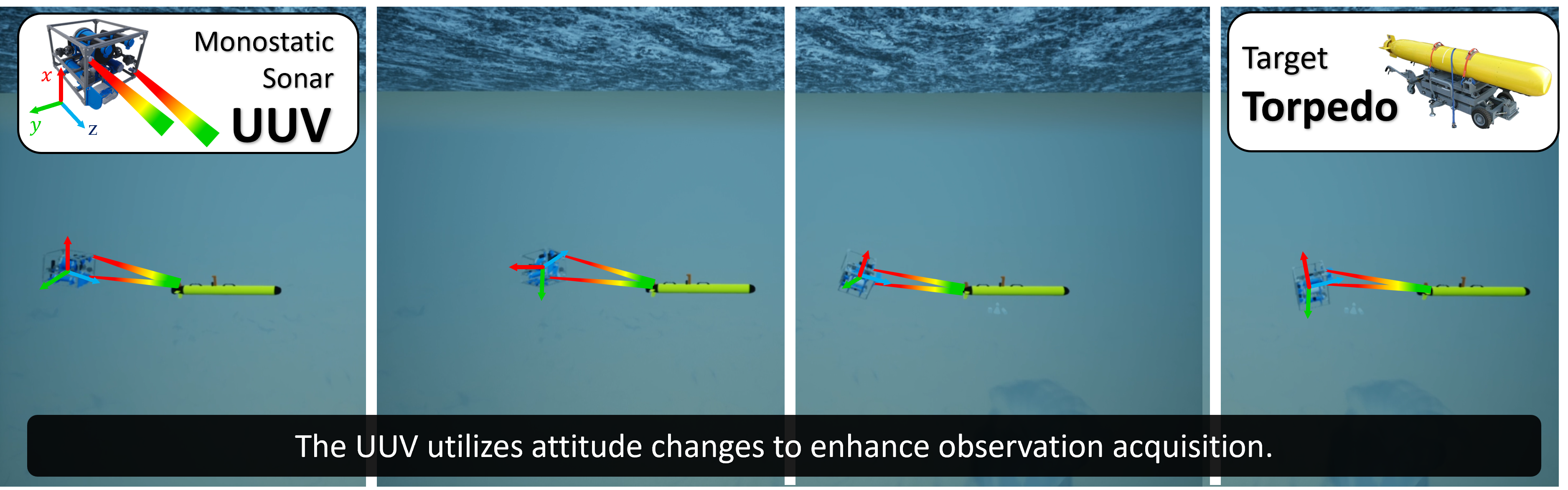}
  \vspace{-20pt}
  \caption{Snapshots from simulation experiments.}
  \label{simulator_results}
\end{figure*} 

Through MATLAB simulation, the results shown in Fig. \ref{fig:error} and  Fig. \ref{fig:trajectory} are obtained. Fig.~\ref{fig:error} consists of two subplots, illustrating the temporal evolution of estimation error (red) and tracking error (blue). The errors are initially large, rapidly decrease, and then remain within a small fluctuation range in the steady-state phase. A zoomed-in view of the critical time interval ($k \approx 200$ to $250$) is provided to highlight finer details. This figure demonstrates that the proposed algorithm effectively reduces errors and maintains stability, validating the effectiveness of the designed estimator and controller. Fig. \ref{fig:trajectory} illustrates the trajectory of a UUV in three-dimensional space and its relative positioning to the target. The red curve represents the target's trajectory, while red triangles indicate the target’s real-time positions in the world frame. The blue and green star markers correspond to measurements from sensors $s1$ and $s2$, respectively. A zoomed-in window highlights the UUV’s attitude changes in the world frame, providing a clearer view of its motion details. This figure demonstrates that compensating observations through attitude adjustments is an effective approach to ensuring accurate relative state estimation.

\textbf{Simulator test:}
The simulator test is conducted using HoloOcean \cite{potokar2022holoocean}, an open-source underwater simulator built on Unreal Engine 4, which provides a realistic environment for testing underwater robotic algorithms. The UUV used in this test is the Hovering UUV, a custom in-house hovering autonomous underwater vehicle designed for high maneuverability and precise control.

\begin{figure}
\centering
  \includegraphics[width=8cm]{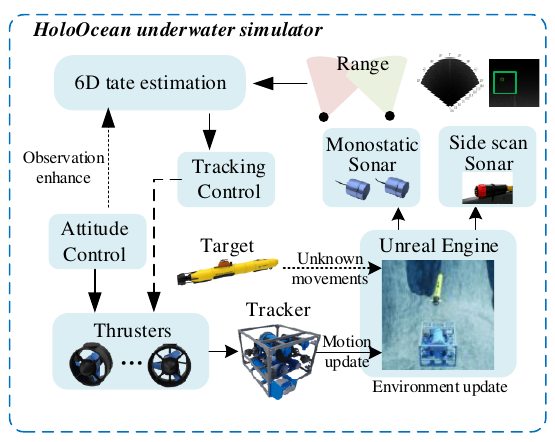}
 \caption{UUV simulator test setup}
  \label{fig:test_setup}
\end{figure}

As shown in Fig. \ref{fig:uuv}, the Hovering UUV is equipped with 8 thrusters, enabling full 6-DOF control. This allows the UUV to maintain both position and attitude stability in dynamic underwater environments. The simulator test setup is described in Fig. \ref{fig:test_setup}. The monostatic sonar and SSS acquire target distance information, which is processed by the 6D state estimation module to compute the target's relative position and velocity. Note that here, we assume the monostatic sonar is equipped on a gimbal.
Attitude control adjusts the UUV's direction via thrusters to optimize sensor observation and compensate for measurement errors. The tracking control dynamically adjusts the UUV's trajectory based on the estimated target state, ensuring precise and stable target tracking.

Fig. \ref{simulator_results} presents snapshots from simulation experiments. The results demonstrate that the UUV can achieve persistent state estimation and target tracking by compensating observations through attitude adjustments. More simulation and experimental details can be found in the URL \url{https://youtu.be/aiITn9bqdzk}.

\section{CONCLUSIONS}
In this work, we have proposed a novel persistent relative 6D state estimation framework for a single UUV to track an uncooperative target using only two noisy range measurements. By introducing an observability-enhanced attitude control strategy and a Lyapunov-based tracking control, we have improved estimation accuracy and long-term stability. Theoretical analysis and simulations have demonstrated the robustness of our approach, providing a scalable, infrastructure-free solution for underwater target tracking. In future work, we aim to further optimize the attitude control algorithm of UUV, enabling it to adaptively adjust its orientation based on the target’s motion to maximize observability.

%%%%%%%%%%%%%%%%%%%%%%%%%%%%%%%%%%%%%%%%%%%%%%%%%%%%%%%%%%%%%%%%%%%%%%%%%%%%%%%%
%\balance
\bibliographystyle{IEEEtran}
\bibliography{IEEEfull}
% \bibliographystyle{IEEEtran}
% \bibliography{IEEEtran}

\end{document}